\documentclass[10pt, a4paper]{article}

\usepackage{lrec-coling2024} % this is the new style
\usepackage{annotates}

\usepackage{microtype}
\usepackage{booktabs}
\usepackage{CJK}
\usepackage{amsmath}
\usepackage{multirow}
\usepackage{enumitem}
\usepackage{caption}
\usepackage{placeins}
\usepackage{xurl}

\title{Multi-Dimensional Machine Translation Evaluation: \\Model Evaluation and Resource for Korean}

\name{Dojun Park, Sebastian Padó} 
\address{IMS, University of Stuttgart, Germany \\\texttt{\{dojun.park,sebastian.pado\}@ims.uni-stuttgart.de}}

       \abstract{Almost all frameworks for the manual or automatic
         evaluation of machine translation characterize the quality of
         an MT output with a single number. An exception is the
         Multidimensional Quality Metrics (MQM) framework which offers
         a fine-grained ontology of quality dimensions for scoring
         (such as style, fluency, accuracy, and terminology). Previous
         studies have demonstrated the feasibility of MQM annotation
         but there are, to our knowledge, no computational models that
         predict MQM scores for novel texts, due to a lack of
         resources.
         In this paper, we address these shortcomings by (a) providing
         a 1200-sentence MQM evaluation benchmark for the language
         pair English--Korean and 
         (b) reframing MT evaluation as the multi-task problem of  simultaneously predicting several MQM scores using SOTA language models, both in a reference-based MT evaluation setup and a reference-free quality estimation (QE) setup. 
         We find that reference-free setup outperforms its counterpart
         in the style dimension while reference-based models retain an
         edge regarding accuracy. Overall, RemBERT emerges as the most
         promising model. %
         Through our evaluation, we offer an insight into the
         translation quality in a more fine-grained, interpretable
         manner. \\ \newline }

\begin{document} 

\maketitleabstract

%% ---------------1. Introduction----------------
\section{Introduction}

%\spa{This formulation will antagonize people working on MT eval -- avoid. Maybe say the focus has been on simplifying MT eval into a single number, but that that's oversimplifying things}
%\spa{I'd actually think about introducing a small table that shows
%  (made up toy) examples of cases where the accuracy is good but the style is %not appropriate etc. That would be a nice motivation.}

%\dpa{I've edited (b) in the abstract with an emphasis on multi-task learning.}

Machine Translation (MT) evaluation refers to assessing the quality of translations generated by MT systems. Since the early stages of machine translation, evaluating its output has been an integral concern, as it enables systems to be assessed, refined, and enhanced \citep{Dorr2011Chapter5M}. %\dpa{would the citation help to support the long concern of MT evaluation along with MT?}
As recent transformative advances in MT technologies \citep{Bahdanau2014NeuralMT, Wu2016GooglesNM, vaswani2017attention} enabled higher-quality, more nuanced translations, the importance of MT evaluation has evolved correspondingly, requiring a more sophisticated ability to measure quality accurately.
%\spa{Why has the importance of MT eval grown with recent models? Not clear. Do you mean because the models have become much better so simple approaches don't cut it any more, or something else?}
%\dpa{Yes, I wanted to say that, since the MT models produce more nuanced translations including idiomatic and context-dependent expressions, MT evaluation needed to be more sophisticated correspondingly to capture those advanced translations. I made a revision in the paragraph. Do you think my intention in this way suits the context?}

While recent advancements by neural metrics
\citep{Zhang*2020BERTScore:, sellam-etal-2020-bleurt,
  rei-etal-2020-comet} have significantly brought the evaluation of
machine translation forwards, virtually all current work on automatic
MT evaluation distills the complexity of translation quality into a
single score to be annotated and predicted, respectively. A
single-score approach, while greatly simplifying computational
modeling, arguably falls short of capturing the inherent
multidimensional concept of translation quality. This limitation
underscores the importance of a fine-grained evaluation.  As an
illustration, consider the three types of translation errors in
Table~\ref{tab:trans-ex}: \textit{Accuracy} errors fail to convey the
meaning of the input. \textit{Fluency} errors arise from outputs that
fail to be grammatical and natural. \textit{Style} errors finally
 change the style of the input substantially.

Being able to distinguish different aspects of translation quality
allows for a fairer comparison across different MT systems in a line
and reveals their strengths and weaknesses across different aspects of
evaluation criteria \citep{avramidis-etal-2018-fine,
  Klubicka2018QuantitativeFH}. Interestingly, early work on MT
evaluation generally distinguished two major aspects of translation
quality, namely adequacy and fluency
\citep{white-oconnell-1993-evaluation}. However, the field later saw a
focus on single-score evaluations that arose both from simpler
modeling \citep{papineni-etal-2002-bleu} and from concerns about
annotation reliability \cite{chatzikoumi_2020}.

\begin{table}[t!]
%\resizebox{\columnwidth}{!}{%
\begin{tabular}{lp{5.1cm}}
\toprule
\textbf{Error Type} & \multicolumn{1}{c}{\textbf{Sentence}} \\ \midrule
\textit{Original}    & \textit{The cat chased the mouse.}             \\
Accuracy    & The cat chased the \textbf{ball}.              \\
Fluency     & The cat the mouse \textbf{chased}.             \\
Style       & The cat \textbf{found itself in pursuit of} the mouse.      \\ \bottomrule
\end{tabular}%
%}
\caption{Different types of translation errors}
\label{tab:trans-ex}
\end{table}

An MT evaluation framework that takes these distinctions seriously is
MQM, the \textit{multidimensional quality metrics} framework
\cite{Lommel2014MultidimensionalQM}. It decomposes translation quality
into a number of aspects and their subaspects. Errors are identified
according to their error types and severity levels and they are
converted into numerical scores by their pre-defined weights (see
Section~\ref{sec:mult-qual-metr} for details).  MQM is a robust
 scheme that corresponds well to judges' overall assessment
of translation quality and provides nuanced insights into the
properties of MT output \cite{Freitag2021ExpertsEA}. However, there
are few corpora annotated with MQM scores, and correspondingly, little
computational work that assesses the automatic prediction of MQM
scores for the purposes of MT evaluation.

In this paper, we show that a suitably adapted version of MQM lends
itself to comparatively easy modeling on top of current neural
language models which makes use of three key error dimensions:
accuracy, fluency, and style. This multidimensionality is not a huge
concern any more, since these models straightforwardly support multi-task learning.

Our study provides three main contributions: First, we present a
resource of MQM-annotated dataset for English-Korean translation
evaluation, a language pair known to be challenging both for
translation and for evaluation
\cite{choi-etal-2018-automatic}. Second, we train and evaluate models
for automatic MQM score prediction. Third, we identify the optimal
conditions for our method by scrutinizing an array of models on varied
data scales and diverse inputs, showing that robust prediction is
possible even with relatively limited training data. The corpus and
the model code are publicly available on our GitHub repository at
\url{https://github.com/DojunPark/multidimensional_MTE}.
%\spa{Can
%  you de-anonymize the repo, or is the server only for anonymous
%  repos? In that case, we might want to move to a non-anonymous repo.}
%  \dpa{I've updated the link to refer to my original github repository, based on which the anonymous repo was created. I will update the training code as soon as we are done with the paper revision!}

\paragraph{Plan of the paper.}
Section 2 introduces related work. Section 3 describes our MQM resource for the the language pair of English–Korean. Section 4 discusses the model architectures and the experimental setup. Section 5 presents our experiments along with their results. Finally, Section 6 discusses  our findings and directions for future research.

%\spa{leaves comments like this}
%\dpa{leaves comments like this}

%% ---------------2. Related Work----------------
\section{Related Work}

% \spa{The contents here are OK but you need to compress everything into at most one page. The things that are \textit{directly} relevant get a short paragraph; aspects that should be mentioned but are not directly relevant get a sentence or just a reference. For example, the difference between symbolic and neural MT Eval is not super relevant for this paper given that most of them predict a single number, which is our main point.}
% \spa{BUT you should definitely extend the MQM paragraph. Imagine that
%   a reader is not familiar with MQM, what do they need to know to
%   understand the rest of the paper? This requires a fairly detailed
%   discussion of the motivation and setup of MQM, including a concrete
%   list of error types (table / figure) and a concrete example of a
%   scored translation (maybe this can integrated into the ``example''
%   table in the Intro, though.)  This is super relevant for the rest of
%   the paper because we build on it, so can reasonably take up up to
%   half a page.}
% \spa{An alternative would be to remove MQM from the Related Work section and
% make it part of the Resource section. Maybe it would be even better, because then Related Work would be less asymmetrical; but I think both strategies might work and will leave the decision to you.}
% \dpa{I see, I will locate the MQM part to the resource section.}

\subsection{Human Evaluation}

% \subsubsection{Traditional Metrics}
Adequacy and fluency \citep{white-oconnell-1993-evaluation} are among the most traditional approaches to human evaluation. Adequacy assesses if translations by MT systems convey the meaning of the source text accurately, while fluency evaluates the naturalness and fluency of the target text, paying particular attention to grammar and idiomatic expressions.

Ranking \citep{Duh2008RankingVR} is another traditional method where translations from different systems are compared and ranked sentence by sentence. This relative ranking method often yields better inter-annotator agreement than evaluations based on adequacy and fluency \citep{koehn2010statistical}.

\subsection{Automatic Evaluation}

\paragraph{String-based Metrics.}
BLEU \citep{papineni-etal-2002-bleu} is the most common automatic
metric for MT quality evaluation. It assesses machine-generated
translations by computing n-gram precisions, which focus on precision,
and imposing a brevity penalty, which serves to capture the aspect of
recall. While being pointed out for its limitations
repeatedly \citep{marie-etal-2021-scientific, Chauhan2022ACS}, BLEU
still remains widely used in a majority of MT publications
\citep{marie-etal-2021-scientific, Chauhan2022ACS}. METEOR
\citep{banerjee-lavie-2005-meteor} addresses some limitations of BLEU
by incorporating stemming and synonymy.
%, at the cost of increased
%computation.

Instead of focusing on the word level n-gram, Character F-score (ChrF) \citep{popovic-2015-chrf} evaluates translations based on character-level n-gram overlaps. An enhanced version, ChrF++ \citep{popovic-2017-chrf}, extends the original metric by also considering word-level n-grams in its evaluation and they are often considered as alternative metrics to BLEU.

Translation Error Rate (TER) \citep{snover-etal-2006-study} is another automatic metric based on the edit distance such as insertion, deletion, substitution, and shift. HTER (Human-mediated Translation Error Rate) integrates human intervention in the evaluation process by comparing the machine output to a version post-edited by a human translator.

\begin{table*}[t]
  \centering
  \small
  \setlength{\tabcolsep}{3pt}
%\resizebox{\textwidth}{!}{% 
\begin{tabular}{ll|l}
\hline
Major Cat. & Minor Cat.                      & \multicolumn{1}{c}{Description}                                     \\ \hline
\multirow{4}{*}{Accuracy}    & Addition                  & Translation includes information not present in the source.         \\
                             & Omission                  & Translation is missing content from the source.                     \\
                             & Mistranslation            & Translation does not accurately represent the source.               \\
                             & Untranslated text         & Source text has been left untranslated.                             \\ \hline
\multirow{6}{*}{Fluency}     & Punctuation               & Incorrect punctuation (for locale or style).                        \\
                             & Spelling                  & Incorrect spelling or capitalization.                               \\
                             & Grammar                   & Problems with grammar, other than orthography.                      \\
                             & Register                  & Wrong grammatical register (eg, inappropriately informal pronouns). \\
                             & Inconsistency             & Internal inconsistency (not related to terminology).                \\
                             & Character encoding        & Characters are garbled due to incorrect encoding.                   \\ \hline
\multirow{2}{*}{Terminology} & Inappropriate for context & Terminology is non-standard or does not fit context.                \\
                             & Inconsistent use          & Terminology is used inconsistently.                                 \\ \hline
Style                        & Awkward                   & Translation has stylistic problems.                                 \\ \hline
\multirow{6}{*}{\begin{tabular}[c]{@{}l@{}}Locale \\ convention\end{tabular}}& Address format            & Wrong format for addresses.                                         \\
                             & Currency format           & Wrong format for currency.                                          \\
                             & Date format               & Wrong format for dates.                                             \\
                             & Name format               & Wrong format for names.                                             \\
                             & Telephone format          & Wrong format for telephone numbers.                                 \\
                             & Time format               & Wrong format for time expressions.                                  \\ \hline
\multicolumn{2}{l|}{Other}                               & Any other issues.                                                   \\ \hline
\multicolumn{2}{l|}{Source error}                        & An error in the source.                                             \\ \hline
\multicolumn{2}{l|}{Non-translation}                     & Impossible to reliably characterize distinct errors.                \\ \hline
\end{tabular}%
%}
\caption{MQM hierarchy from \citet{Freitag2021ExpertsEA}}
\label{tab:mqm}
\end{table*}

\paragraph{Neural Metrics.}
Neural metrics represent an advanced approach for automatic evaluation of machine-generated translations. It can be categorized as either unsupervised or supervised \citep{math11041006}. Unsupervised metrics such as BERTscore \citep{Zhang*2020BERTScore:} and YiSi \citep{lo-2019-yisi} leverage the contextual embeddings of pre-trained models to measure the semantic similarity between words in candidate and reference translations, enabling a more informative evaluation. 

Supervised metrics such as BLEURT \citep{sellam-etal-2020-bleurt} and COMET \citep{rei-etal-2020-comet} go a step further. They fine-tune pre-trained models on manually scored data sets utilizing specific human evaluation metrics, thereby producing assessments that more closely approximate human judgments. Recent studies have confirmed superior performance of supervised neural metrics in correlating with human assessments, outperforming other traditional and unsupervised neural metrics \citep{mathur-etal-2020-results, freitag-etal-2021-results}.

However, there are potential pitfalls: These metrics are prone to a
higher susceptibility to overfitting on the data they are trained
on. Also, if the embeddings used in both the MT and evaluation models
are too similar, there is a risk of bias, potentially
skewing evaluation outcomes.

\subsection{Quality Estimation}
Quality Estimation (QE) is considered an alternative approach to MT evaluation. As a reference-free evaluation, it assesses the quality of a translation by taking only the source sentence and its machine-generated translation without a reference translation \citep{specia2018quality}. This technique is advantageous as it provides an estimation of translation quality without requiring reference translations, which might not always be available. 

QuEst \citep{specia-etal-2013-quest} and QuEst++
\citep{specia-etal-2015-multi} are statistical QE systems which
operate by leveraging linguistic features from both the source
sentence and its translation and learning to predict translation
quality based on these features.

Following the general trend, neural methods have recently become
central in QE systems. OpenKiwi \citep{kepler-etal-2019-openkiwi}
incorporates leading QE systems from the WMT 2015–18 tasks into its framework.
%\spa{'incorporates' into what? Below you also talk about the 'predictor-estimator architecture' but that was never discussed. Maybe OpenKiwi needs another sentence or two?} 
%\dpa{By "incorporate", I meant that it has four different QE systems in it, which have been the winners from WMT 2015–18 QE campaigns. I've added additional information for predictor-estimator architecture, with a direct reference to its original paper.}
TransQuest\citep{ranasinghe-etal-2020-transquest}, building on XLM-RoBERTa
embeddings \citep{Conneau2019UnsupervisedCR}, has surpassed other
state-of-the-art systems \citep{specia-etal-2020-findings-wmt}.
Most recently, COMETKiwi \citep{rei-etal-2022-cometkiwi} achieved a
new state-of-the-art in the WMT 2022 shared task on QE
\citep{zerva-etal-2022-findings}. It has brought QE performance forward by integrating the COMET framework \citep{rei-etal-2020-comet} with the predictor-estimator architecture \citep{Kim2017PredictorEstimatorUM}, where a predictor processes both source and target sentences to predict target words, and an estimator uses these feature vectors to estimate translation quality. This advancement has further blurred the lines between traditional MT evaluation and Quality Estimation.

%% ------------3. Resource-----------------
\section{An MQM Resource for the Language Pair English--Korean}
\label{sec:an-mqm-resource}

\subsection{Multidimensional Quality Metrics}
\label{sec:mult-qual-metr}

MQM or Multidimensional Quality Metrics 
\citep{Lommel2014MultidimensionalQM} is a framework specifically
designed to identify translation issues and transform them into
quantifiable scores. It features a hierarchical structure of
translation error types as shown in Table \ref{tab:mqm}.
%\spa{Show the hierarchy, or an excerpt of
%  them, here, to give people an idea of what MQM looks like in
%  practice. You can copy a figure from an MQM paper if you find a
%  suitable one, provided that you give credit to the original publication
%  (``figure from [CITE]'').}
%  \dpa{Yes, I brought the MQM hierachy table from Freitag et al., 2021a.}
%, grouped into categories such as accuracy,
%fluency, and style among others.
In addition, MQM proposes to weigh each error by its severity. It
offers three weights: minor error (weight 1), major error (weight 5),
and critical error (weight 25). At the segment level, we can aggregate
errors to obtain first category-specific scores, and then aggregate
category scores to obtain overall scores.

MQM sees itself as a general framework whose users should select the
most pertinent error categories and severities for each particular
translation context. We now describe our use case
and then proceed to describing our adaptations to MQM.
%\spa{What I'm trying to do here is to separate cleanly what MQM in
%  general says (this subsection) and how we adapt it to our use case %(subsection 3.3).}

\subsection{Constructing an English--Korean Parallel Dataset}

To construct our dataset, we start with parallel corpora, then
generate translations and quality assessments using a two-step
approach: paraphrasing and quality evaluation.

%\spa{Please add
%  one sentence each for each the corpora to describe where they come
%  from and what they are about. Please also add refs or URLs} \dpa{I added the %description for each corpus by mentioning their source}
From the OPUS (Open Parallel Corpus) project \citep{TIEDEMANN12.463}, we chose two English-Korean parallel corpora to capture diverse linguistic styles: Global Voices \citeplanguageresource{TIEDEMANN12.463resource} and TED Talks 2020  \citeplanguageresource{Reimers2020MakingMSresource}. Global Voices consists of news articles crawled from its website in 46 languages, while TED Talks 2020 features around 4,000 transcripts from TED and TED-X events as of July 2020, covering 108 languages.
  From
each, we randomly sampled 600 translation pairs, summing up to
1200. Out of these, 1000 pairs form our training set, with the
remaining 200 equally divided as validation and test set. This makes
our dataset comparable in size to COMET, which employed 1000
translation pairs for each language pair \citep{rei-etal-2020-comet}.

\paragraph{Paraphrasing.}
%\spa{MISSING: rationale for paraphrasing!!}
%\dpa{I added the reason for it}
Parallel corpora are often used as gold standard in
translation. When such corpora only contain a single manual
translation, however, only capture a small part of the space of
translations, and towards correctness. We believe that we can enhance
the robustness of our evaluation (and evaluation models) by
considering a larger sample of potential translations, including
somewhat flawed ones which exhibit a range of natural errors.

To do so, we automatically paraphrase our parallel corpus. We opt for
proprietary online services for both paraphrasing the English source
sentences and subsequently translating the paraphrased content into
Korean, given their notable superiority over open-source alternatives.
We obtained paraphrases from ChatGPT gpt-3.5-turbo\footnote{\url{https://platform.openai.com/docs/models}} using the following prompt:
%\spa{Is there a reference we can list?} 
%\dpa{Unfortunately, I couldn't find a relevant publication for it. What about referring to its official API document as a footnote instead? Or, could we possibly cite it as a reference this way?} \citep{gpt-3.5}, we
%issued the following prompt:\spa{If we need a bit more space, we can move
%the paraphrasing description to the appendix.}
%\dpa{Okay!}

\begin{verbatim}
Please rewrite the given sentence in
English while maintaining the same
meaning, using different vocabulary
or sentence structures:
[English sentence]
\end{verbatim}

The placeholder ``[English sentence]'' is replaced with the actual sentence for paraphrasing, then translated into Korean using Google Translate.

\begin{table}[t]
  \centering
%  \setlength{\tabcolsep}{5pt}
%  \resizebox{\columnwidth}{!}{%
%  \small
\begin{tabular}{lp{1.8cm}p{2.2cm}}
\toprule
                 & \textbf{Global Voices} & \textbf{TED Talks 2020}          \\ \midrule
Genre            & News          & Presentation Transcript \\
Style            & Formal        & Conversational          \\
Total Pairs      & 9,381         & 399,413                  \\
Sampled Pairs    & 600           & 600                     \\
Avg. Len. (src.) & 17.74         & 15.99                   \\
Avg. Len. (ref.) & 12.90         & 10.98                   \\
Avg. Len. (tgt.) & 13.02         & 11.23                   \\ \bottomrule
\end{tabular}%
%}
\caption{Corpus summary statistics}
\label{tab:stats}
\end{table}

Table \ref{tab:stats} shows statistics of the selected parallel corpora and the generated sentences. Global Voices is news-based and formal, while TED Talks 2020 contains conversational presentation transcripts. Notably, the generated Korean sentences are slightly longer than their corresponding reference sentences. The increase in sentence length suggests that the paraphrasing step introduces structural and stylistic variation, which could impact the quality of translation.
%\spa{Say something about the relevance?}
%\dpa{I've added an argument supporting a possible variation of quality that may have been introduced in paraphrasing step.}

% \dpa{perhaps we could bring this table to appendix or just completely remove?}
% \spa{In principle yes -- but I'm not sure we need to.}
% \begin{table*}[t]
% \centering
% \begin{tabular}{@{}cl@{}}
% \toprule
% \textbf{Step} & \multicolumn{1}{c}{\textbf{Sentence}}           \\ \midrule
% Original      & But I didn't tell my parents.                   \\
% Paraphrased   & However, I refrained from informing my parents. \\
% Translated    & \begin{CJK}{UTF8}{mj}그러나 나는 부모님께 알리는 것을 자제했다.\end{CJK}                        \\ \bottomrule
% \end{tabular}
% \caption{Example of the translation generation process}
% \label{tab:gen_process}
% \end{table*}

% Table \ref{tab:gen_process} offers a sample from our translation generation process, highlighting the progression from the original English sentence, through paraphrasing, to the final Korean translation. This systematic procedure enabled us to generate Korean sentences suitable for subsequent quality assessments.

\subsection{Annotation of Annotation Quality}

For our annotation, we make the following adaptations to the general
MQM framework:

\begin{itemize}
\item Error Dimensions: We select three major error dimensions of
  accuracy, fluency and style. We leave out the major category of
  terminology since our selected corpora comprise general
  texts where terminology is not a major factor. We also exclude
  the dimensions of audience appropriateness and design and markup 
  since our corpora neither target a specific audience nor include 
  graphical presentations.
  %\spa{Also rationale for
  %  leaving out more fine-grained categories!}
  %  \dpa{I added the reason for it}

\item Sub-error Type Integration: Given the infrequency of formatting 
  issues in our evaluation, we chose to integrate them as a sub-error 
  type under the fluency dimension, deeming it unnecessary to maintain 
  them as a primary error dimensions. Furthermore, since untranslated text --- initially a sub-error type under 
  accuracy --- impedes both the original meaning and the readability, we 
  include it under both the accuracy and fluency dimensions.
  %\spa{Not sure I get that.}
  %\dpa{what about like this?}

% \item Stylistic Considerations: Given that stylistic alterations often
%   span beyond clauses or phrases, we approached style errors
%   differently. Rather than counting them per word, we identified them
%   per text unit, which refers to any continuous string of words
%   perceived as erroneous.
%    \spa{Above I say MQM generally counts per
%    segment. The formalization doesn't make explicit whether it counts
%    at word, sent, segment level. Either keep this underspecified and
%    delete this point, or make the descriptions more specific.}
%    \dpa{I think we can leave out this since these details are anyway
%    included in the guidelines!}

\item Error Severity Classification: We distinguish two severity
  levels: major errors and minor errors. We omit the
  critical error category due to its inherent subjectivity, as
  supported by extant literature \citep{Freitag2021ExpertsEA}.
\end{itemize} 
%
%\spa{If we run short of space,
%  we might want to delete this formalization; it's very simple and
%  doesn't really add much to the description above.} \dpa{I see!}
We calculate the scores for the three dimensions accuracy \( S_a \),
fluency \( S_f \), and style \( S_s \) as follows:
\begin{align}
S_a &= 5 \times E_{a,m} + 1 \times E_{a,i},\\
S_f &= 5 \times E_{f,m} + 1 \times E_{f,i},\\
S_s &= 5 \times E_{s,m} + 1 \times E_{s,i},
\end{align}
where each score is the sum of the products of their respective major
errors (weighted by 5) and minor errors (weighted by 1). Specifically,
for each dimension \( d \) (accuracy, fluency, style), the score
\( S_d \) is determined by the major \( E_{d,m} \) and minor
\( E_{d,i} \) errors of that dimension.\ The total score,
\( S_{\text{total}} \), is:
\begin{equation}
S_{\text{total}} = S_a + S_f + S_s,
\end{equation}
%
%\spa{Just to clarify: so there are NO manual annotations of overall
%  annotation quality, they are computed from the manual annotations of
%  the three individual dimensions -- is that correct?}
%\dpa{Yes, the total score is derived from the annotations for the three 
%dimensional scores.}
%
We created a set of guidelines for our EN-to-KO translation evaluation
based on the MQM guidelines\footnote{\url{https://themqm.org/}}, given in
Appendix \ref{sec:annot-guid}.
%  , given in Appendix\
%\ref{sec:annot-guid}.
%
The annotation was carried out primarily by an evaluator proficient in
English at a level above CEFR C1
\footnote{\url{https://www.coe.int/en/web/common-european-framework-reference-languages/table-1-cefr-3.3-common-reference-levels-global-scale}}
and native in Korean with a background in computational linguistics.
The five most frequent error types annotated are: mistranslation;
unnaturalness; structure; untranslated text; omission. Details on
error types and score distributions are given in Appendix \ref{sec:score-error-distr}.

%Appendix\
% provides histograms of the score
%distributions and specific error types annotated.

The annotation was timed at approximately 5 min/unit. While
this may sound long, this is comparable to a previous annotation study
by \citep{Mariana2014TheMQ} who report an annotation speed of around
10 min/unit, albeit for longer sentences.

%\spa{Would it be possible to list average scores for the three dimensions, %maybe separately for the original and paraphrased sentences?}
%\dpa{Actually I only carried out the annotation on the paraphrased sentences. %If you mean those scores, the average scores of the paraphrased ones are: %
%Accuracy: 13.24
%Fluency: 4.23
%Style: 2.42}

%\spa{Can you please estimate the speed of annotation, and compare it
%  to other annotation projects in MT eval or MQM? (Provided there are
%  publicly available numbers.)}

\subsection{Validation of MQM Scores}

We can now ask two questions: (a), are the MQM scores
that we have obtained \textit{reliable}? (b), do they provide us with
\textit{additional} information compared to single-score metrics, as
we claimed above?

To address point (a), we employ cross-validation with two independent
annotators, namely two undergraduates who satisfied the same
prerequisites as our primary annotator. They independently annotated a
subset (100 translation units) of our primary data using the
guidelines we established. 
%\spa{Did their speed differ considerably
% from the main annotator?} \dpa{The first annotator took 3.3 min/unit and the second one 5.22 min/unit, which is, I assume, similar to the amount of time I spent.}
We use Kendall's Tau correlation as our
evaluation metric (see Section~\ref{sec:exp-setup} for details).

%Given that parallel corpora typically represent the gold
%standard in translation, our paraphrasing step introduces essential
%variations for robust model training.

\begin{table}[t]
\centering
  \begin{tabular}{@{}ccc@{}}
\toprule
  Accuracy & Fluency & Style \\
  \midrule
  0.54 & 0.57 & 0.34 \\
  \bottomrule
  \end{tabular}
\caption{Correlations (Kendall's $\tau$) between annotators by dimension.}
\label{tab:CV}
\end{table}

\begin{table}[t]
  \setlength{\tabcolsep}{4pt}
\begin{tabular}{cccccc}
\toprule
  & Accuracy         & Fluency & Style       & &BLEU \\
  \midrule
Accuracy           & 1                &         &             & &0.17***  \\
Fluency            & 0.29*** & 1       &             & &0.15*** \\
  Style              & 0.10*** & 0.01 & 1 && 0.08*** \\
  \bottomrule
  \end{tabular}
  \caption{Correlations (Kendall's $\tau$) among MQM dimensions and between MQM dimensions and BLEU score (stars indicate statistical significance).}
  \label{tab:MQM-corr}
\end{table}

Table \ref{tab:CV} shows the correlation between our primary dataset
scores and average scores from two cross-validators. Fluency
correlates highest at 0.57, with accuracy at 0.54, and style at 0.34,
suggesting that evaluating translation style may be more
subjective. Yet, the cross-validation confirms a robust level of
agreement with our primary evaluation.

Regarding point (b), Table\ \ref{tab:MQM-corr} shows that there are
some correlations among the three MQM dimensions, but they are
sufficiently mild to warrant the conclusion that the scores indeed
measure distinct aspects of quality. While the correlations for style
are generally low, we find a correlation of about 0.3 between accuracy
and fluency, which we take to reflect a general cline between `bad'
and `good' translations and which is presumably also related to the
presence of untranslated text, the 4-th most frequent error
category. This interpretation is further supported by the pattern of
correlations between MQM scores and (inverse) BLEU scores we obtain
from the SacreBLEU \citep{Post2018ACF} implementation, shown in the
right column: BLEU correlates to a similar extent with accuracy and
fluency, but the correlation overall remains weak (0.15-0.20). We take
this to mean that the MQM scores offer a fine-grained evaluation,
capturing nuances potentially missed by single-score metrics such as
BLEU.

%%%% ------------4. Modeling MQM--------------
\section{Modeling Multidimensional Translation Quality}

We carry out a series of experiments in predicting the manual MQM
translation quality judgments with various language models: Our main
experiment (Experiment 1) assesses the performance of a set of
selected pre-trained models for multi-score prediction.  Experiment 2
assesses the impact of training data size on model
performance. Experiment~3 compares multi-score and single-score
prediction of overall translation quality, and Experiment~4 compares our
models against the COMET model family.

\begin{figure}[tb!]
  \centering
  \includegraphics[width=\linewidth]{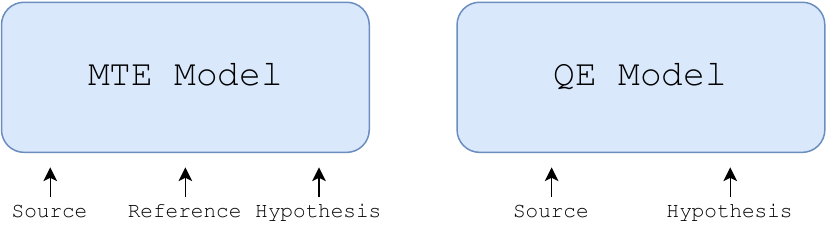}
%  \caption{Input configurations for MTE and QE Models}
%\end{figure}
%\begin{figure}[tb!]
%  \centering
  \hrule\hrule
  \includegraphics[width=\linewidth]{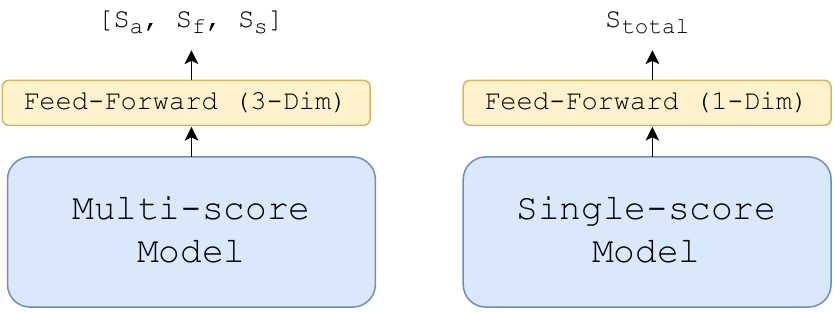}
  \caption{Top: Input configurations for MTE and QE Models. Bottom: Output layer setups for multi-score and single-score models}
  \label{fig:models}
\end{figure}

\subsection{Base Model Selection}

We compare six Transformer-based pre-trained LMs with multilingual
capabilities that cover English and Korean.\footnote{We use the
  Huggingface \texttt{transformers} library.} We consider both
encoder-only and encoder-decoder models. In the latter case, we only
use the encoder part which, given its extensive pre-training,
can potentially function similarly to an encoder-only model in creating
inputs for MT evaluation, a regression task
\cite{kocmi-EtAl:2022:WMT}.
%\spa{Add rationale for this procedure!}\dpa{Yes, I added the reason for the decision. Actually I was trying to find some references for using encoder-decoder model's encoder like encoder-only model, but I couldn't find... do you know any possible reference to cite for our case? The report of WMT22 metrics shared task denotes that the top performer ``METRICX XXL'' fine-tuned mT5-30B, but they didn't publish a paper for it, so I cannot know whether they used only the encoder of it, or in a different way.}
%
Concretely, we use Multilingual BERT base cased
\citep{devlin-etal-2019-bert}, XLM-RoBERTa large
\citep{conneau-etal-2020-unsupervised}, and RemBERT
\citep{Chung2020RethinkingEC} (encoder-only) and Multilingual BART
large-50 \citep{Liu2020MultilingualDP}, Multilingual T5 large
\citep{Xue2020mT5AM} and M2M100 1.2B \citep{Fan2020BeyondEM}
(encoder-decoder). For details see Appendix \ref{sec:model-details}.

\begin{table*}[tb!h]
  \centering
  % \small
  \setlength{\tabcolsep}{4pt}
\begin{tabular}{llcccccccc}
\toprule
&   &   \multicolumn{4}{c}{MTE (with reference)} & \multicolumn{4}{c}{QE (reference-free)} \\
  \cmidrule(r){3-6}
  \cmidrule(l){7-10}
&  \multicolumn{1}{c}{\textbf{Model}} & \textbf{Accuracy} & \textbf{Fluency} & \textbf{Style}  & \textbf{Overall} & \textbf{Accuracy} & \textbf{Fluency} & \textbf{Style}  & \textbf{Overall}   \\ \midrule
&mBERT                          & 0.36            & 0.34           & 0.12          & 0.27          
                           & 0.31            & 0.24           & 0.18          & 0.24          \\
Enc. only&  XLM-R                          & 0.32            & 0.32           & 0.09          & 0.24          
                                                                                              & 0.35            & 0.37           & 0.27          & 0.33          \\
&RemBERT                        & \textbf{0.40}   & 0.38           & 0.26          & 0.35          
                         & 0.35            & \textbf{0.43}  & \textbf{0.33} & \textbf{0.37} \\ \midrule
Enc. of &mBART                          & 0.25            & 0.32           & 0.14          & 0.24          
                           & 0.24            & 0.41           & 0.18          & 0.27          \\
Enc.-Dec. &mT5                            & 0.36            & 0.40           & 0.14          & 0.30          
                             & 0.35            & 0.33           & 0.14          & 0.27          \\
&M2M100                         & 0.34            & 0.39           & 0.11          & 0.28         
                          & 0.36            & 0.36           & 0.18          & 0.29          \\ \bottomrule
\end{tabular}%
\caption{Experiment 1: Kendall's Tau between model predictions and human
  scores across models and evaluation settings. Best result per
  dimension bolded.}
\label{tab:exp1}
\end{table*}

\subsection{Input: With vs. Without Reference}

The top part of Figure \ref{fig:models} shows the input configurations
for our two main setups, namely MTE (Machine Translation Evaluation)
models that take the source, reference, and hypothesis sentences; and
QE (Quality Estimation) models that only take the source and
hypothesis, but no reference. In either case, the input is structured
using special tokens similar to BERT's use of \texttt{[CLS]} and
\texttt{[SEP]}. See Table\ \ref{tab:app:input-encoding}
(Appendix~\ref{sec:model-details}) for details.

%SP: moved to appendix
%For encoder-only models, we prefix the input with the
%\texttt{[CLS]} token and inserted \texttt{[SEP]} tokens between input
%units and at the end, using the models' inherent tokens.
%
%For mBART and M2M100, we utilize language-specific tokens. The
%\texttt{en\_XX} and \texttt{\_\_en\_\_} tokens were prefixed to the
%English source and the \texttt{ko\_KR} and \texttt{\_\_ko\_\_} tokens
%to the Korean reference and hypothesis. Each input sequence ends with
%\texttt{</s>}.
%
%For mT5, which did not employ special tokens during its pre-training
%besides \texttt{</s>}, we adopted a slightly different strategy. We
%introduced \texttt{<init>} token at the start and \texttt{<sep>} token
%between input units to maintain consistency, while the \texttt{</s>}
%token marks the end of the sequence.  

\subsection{Output: Single- vs. Multi-Score Prediction}

The bottom part of Figure\ \ref{fig:models} shows that we further distinguish
between a multi-score setup (predicting several scores) and single-score setup (predicting one score). For both model types, we extract embeddings of the initial token from the output of base model taking advantage
of the input encoding described in the previous paragraph.
%Encoder models mBERT, XLM-R, and RemBERT use the inherent \texttt{[CLS]} token. For mBART and M2M100, language-specific tokens \texttt{en\_XX} and \texttt{\_\_en\_\_} are used, respectively. For mT5, we used the introduced \texttt{<init>} token.\spa{Again, model
%  specifics can be moved to an appendix if necessary}
%
The embedding of the initial token is passed to a simple feed-forward regression layer. For multi-score models, we predict a
vector of length three, representing scores for \textit{accuracy}, \textit{fluency}, and
\textit{style} of the translation. Single-score models output a single scalar
value representing the \textit{overall} quality of the translation.
%\spa{why? dimensions? any nonlinearity?}
%\dpa{I'm sorry for the confusion. I misunderstood the term "pooling layer". I was thinking that this term describes the action of pooling the embedding of the intial token, which isn't true. So, I didn't use a pooling layer indeed, and just the embeddings were directly passed to the feed-forward layer. Thank you for pointing it out!}
%

\subsection{Experimental Setup}
\label{sec:exp-setup}

\paragraph{Dataset.}
We use our created MQM-annotated dataset for the English-Korean
language pair, consisting of 1,200 translation units (cf. Section
\ref{sec:an-mqm-resource}). The dataset is split into training (1,000
units), validation (100 units), and test (100 units) sets, each with
an equal distribution between the Global Voices and Ted Talks 2020
corpora. In Experiment 2, training data was reduced to 200, 400, 600,
and 800 units respectively, maintaining the corpus distribution.

\paragraph{Training Regimen.} All our models are regression models,
trained to minimize a mean squared error objective. Optimization is
carried out with stochastic gradient descent, using the AdamW algorithm \cite{loshchilov2017decoupled}. Our learning rate is 2e-6 for all models except mT5, which is adjusted to 2e-5 to accelerate its relatively slower training compared 
to the others. 
%\spa{There's something wrong here: 2e-5 is *more* than 2e-6 which contradicts the statement in the text. Is 2e-6 an error
%  for 2e-4? That seems to me a more plausible learning rate.}
%\dpa{I'm sorry for the confusion! There has been something wrong in my explanation. The learning rate for mT5 was set to 2e-5 indeed. The initial setup by 2e-6 took it too long to decrease its training loss. So, I decided it to increase the learning rate to speed up its learning and to possibly reach its optimal point within the 100 epochs as other models. Therefore, it didn't "require a slower rate", but it was "due to its slower learning". I've edited the line in the paragraph, hopefully it is read reasonably now.}
We train for 100 epochs 
with a batch size of 8. To ensure robustness in our evaluation, we train 
each model across three separate runs and report averages.
%\spa{early stopping?}
%\dpa{I didn't apply early stopping at the end because by fluctuating dozens of epochs, the models seemed to reach lower validation loss.}
%\spa{How did you choose the learning rates?} \dpa{I came to the learning rate by experiments with mBERT, which was the first model I trained. It seemed to converge around 20 to 30 epochs, but I gave some enough epochs to 100. Then, the validation loss fluctuated going up and down and reached the optimal point around 60 and 70. It worked pretty similarly to other models in that I thought I didn't necessarily need to adjust the learning rate. However, in the case of mT5, the learning curve looked pretty much slower compared to others, and then I adjusted it to the number, where each T5 model reached their optima in 81, 97, 98th epoch for MTE models, and 22, 96, 33th epoch for QE models.}

%\paragraph{Computational Resources}
%We trained all models on Google Colab, leveraging the computational %capabilities of the NVIDIA A100 GPU with 40GB memory.

\paragraph{Evaluation Metric.}
For evaluation, we use Kendall's Tau. It offers a more robust measure
of general rank correlation compared to Pearson, which can be
sensitive to outliers. Compared to Spearman, it is more sensitive to
changes between ranks providing a clearer interpretation of pairs that
are in agreement (concordant) and those that are not (discordant). This
choice aligns with the WMT Metrics Shared Task, which have adopted
Kendall's Tau for evaluating segment-level MQM scores since 2021
\citep{freitag-etal-2021-results, freitag-etal-2022-results}:
\begin{equation}
\tau = \frac{\text{\#concordant pairs} - \text{\#discordant pairs}}{\text{\#concordant pairs} + \text{\#discordant pairs}}
\label{eq:kendall}
\end{equation}
A \( \tau \) value of 1 indicates perfect agreement, -1 complete disagreement, and 0 signifies no correlation.

%%%% ------------5. Experiments and Results----------------- 
\section{Results and Discussion}

\subsection{Experiment 1: Model Comparison}
% \spa{This is already defined above, isn't it?}

\begin{figure*}[tb!]
  \centering
  \includegraphics[width=\linewidth]{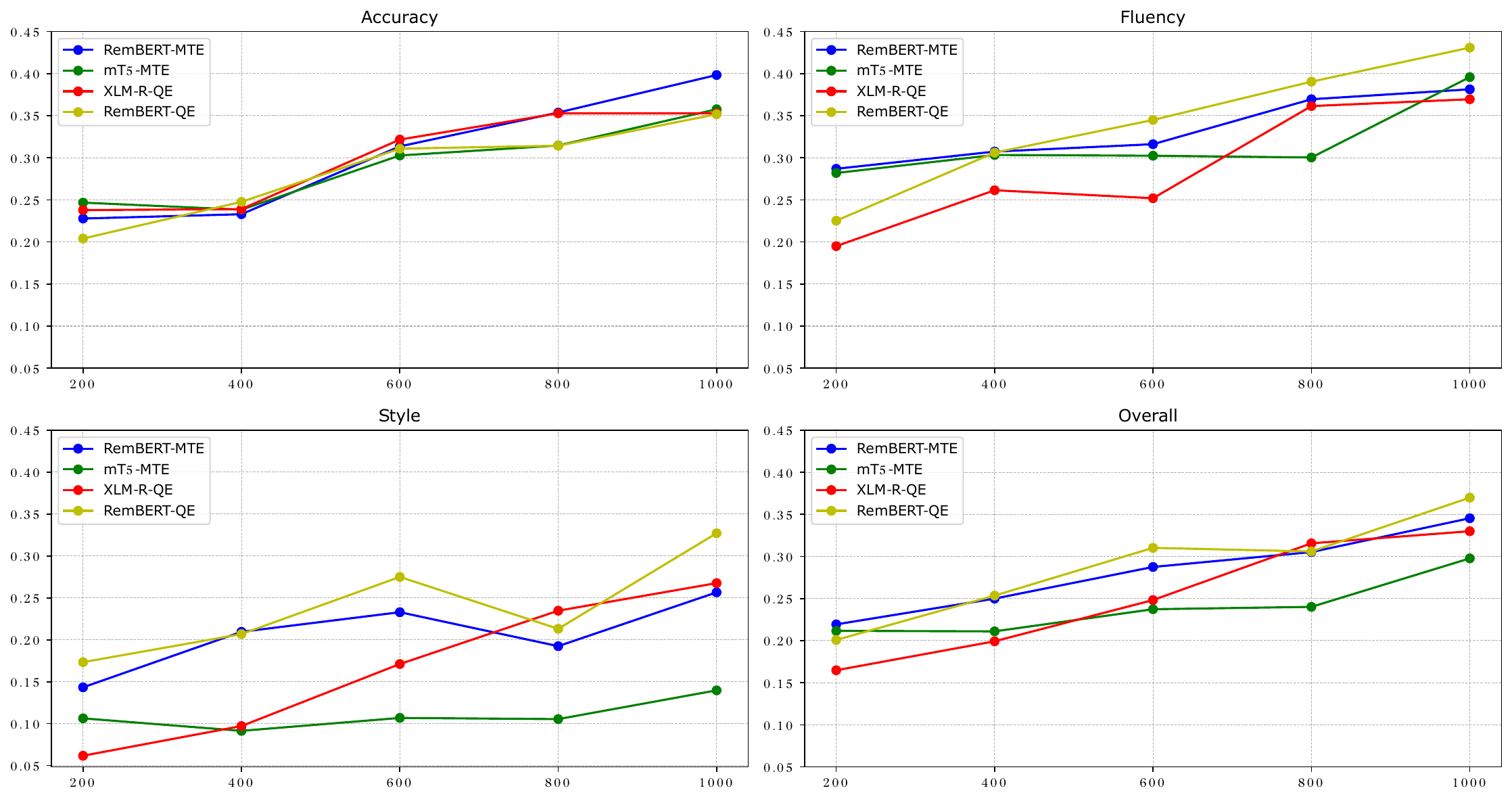}
  \caption{Experiment 2: Kendall's Tau for different amounts of training data (Accuracy, Fluency, Style, and Overall scores).}
  \label{fig:exp2}
\end{figure*}

Table\ \ref{tab:exp1} shows the results for our main experiment,
evaluating a range of language models
% (encoder-only and encoder components of encoder-decoder models)
in both an MLE and a QE setting
(cf. Section~\ref{sec:exp-setup}). Overall, RemBERT stands out in both
MTE and QE setups: RemBERT (QE) achieves the highest performance
overall with an average score of 0.37
and leads in two error dimensions: fluency and style. Following
closely, RemBERT (MTE) yields an average of 0.35 and shows the top
performance in the accuracy dimension. XLM-R (QE) is another
strong performer, with an average score of 0.33. In contrast,
mBART (MTE), XLM-R (MTE) and mBERT (QE) record the lowest average
score with 0.24.
%\spa{Please
%  clarify: the ``average'' row in the Table is NOT the results for
%  predicting overall (average) translation quality, but is the average
%  of the correlations obtained for the individual dimensions. Or is
%  it?  If it is, we should rename it ``Overall (quality)''
%  throughout.}
%  \dpa{no, it differs from the total score (overall quality). I named it as average to differentiate it from the total score that is aimed to represent the overall score. the overall rank is not just the sum of the individual ranks so the average of individual ranks differs from the results of the total scores, which shows the total scores. I intended by the term "average" that it is still focused on dimensional quality by simply showing the average score by individual dimensional scores. What about we add a footnote by the first appearing term "average" that it differs from the total score for better clarification?}

\paragraph{Results by Dimension.} Three of the five best models for
accuracy use the MTE setup. This is to be expected, given that
accuracy concerns the relationship between input and the output of the
MT process. Rather, it is surprising that the best reference-free
model in the QE setting, M2M100 with a score of 0.36, is not far
removed from the best score of any model (0.40).  In contrast, we see
the best results for both fluency and style in the reference-free QE
setting. For fluency, this might be expected since this is primarily a
target language property. It is again surprising for style, though, in
particular given that style shows the largest margin for QE among the
three dimensions. This indicates that the model may currently learn
how well the MT output matches the ``typical'' style of our corpus,
rather than an actual match between the styles of input and output.

\paragraph{Encoder-only Models vs. Encoder Component Models.}
While the top spots are dominated by encoder-only models, they are
followed by four models based on the encoder components of
encoder-decoder models: mT5 (MTE), M2M100 (both MTE and QE) and mBART
(QE) record average scores of 0.30, 0.29, 0.28, and 0.28,
respectively, outperforming three more encoder-only models. This is
particularly true for the accuracy dimension, where M2M100 (QE) and
mT5 (MTE) are close to the best model, RemBERT (MTE), and for the
fluency dimension, where mBART (QE) and mT5 (MTE) are close to RemBERT
(QE). In contrast, the encoder component-based models generally rank
lower in the style dimension. A potential explanation derives from
their distinct training methodologies. Encoder-only models, designed
to predict masked tokens, excel at capturing nuances of word meaning,
while the encoders of encoder-decoder models, optimized to create a
representation of the essence of the source text, might
overlook some nuances of style \citep{Lin2021ASO}.
%\spa{Are there any references to bolster this interpretation? (the tradeoff between encoder only and encoder decoder)}
%\dpa{While I couldn’t find a direct reference supporting our findings on encoder-only and encoder-decoder models, I thought that this survey paper on transformer-based variants could possibly support the internal mechanisms of encoder-only and enc-dec models and why they might have shown those results in our work. Or what about directly referring to Bert and Transformer papers, which illustrate the structural features of those models?}

\paragraph{Error Analysis.} We focus on the two top-performing models,
RemBERT (MTE) and RemBERT (QE). We observe in general that predicted
scores align closely with human assessments when a single type of
error is present and sentences are short. In \textit{mistranslation},
the most commonly identified error, the content domain of the text and
the depth of contextual understanding required play significant roles
in prediction accuracy. Models struggle terms specific to certain
domains such as \textit{color revolution} or \textit{blogosphere}.
Regarding \textit{omission} errors, sentences that leave out frequent
terms like \textit{if} %, \textit{they}, 
and \textit{the} seem to align
more closely with human assessments, while deictic expressions such as
\textit{below} are not only challenging to translate but also to
evaluate. We also found that quality was often overestimated as a
result of minor errors especially when \textit{untranslated} text was
involved -- our models are evidently not always able to detect subtle
discrepancies. See Appendix \ref{sec:score-error-distr} and \ref{sec:error-case} for
details.

%Subtle discrepancies in translation make accurate evaluation more challenging %in distinguishing major and minor errors, which might have been caused to its %insufficient training data by those cases.

%\paragraph{Case Study: Corpus-wise Performance Analysis.}
%As an extension of Experiment 1, we introduce the corpus-wise performance of %the selected models. \dpa{do we actually have a space to add this section as %well as case study at the end?}
%\spa{I think so. Or we can show it as Experiment 4.}
\begin{table*}[tb!h]
\centering
\begin{tabular}{@{}cccccc@{}}
\toprule
\multicolumn{1}{l}{} & \multicolumn{1}{c}{\textbf{Model}} & \multicolumn{1}{c}{\textbf{Accuracy}} & \multicolumn{1}{c}{\textbf{Fluency}} & \multicolumn{1}{c}{\textbf{Style}} & \multicolumn{1}{c}{\textbf{Overall}} \\ \midrule
\multirow{2}{*}{MTE} & RemBERT & 0.40 & 0.38 & 0.26 & 0.35 \\
 & COMET-22 & \textbf{0.30} & 0.10 & 0.02 & 0.28 \\ \midrule
\multirow{2}{*}{QE} & RemBERT & 0.35 & 0.43 & 0.33 & \textbf{0.37} \\
 & CometKiwi & \textbf{0.40} & 0.09 & 0.17 & \textbf{0.37} \\ \bottomrule
\end{tabular}
\caption{Experiment 4: Kendall's Tau correlations of different models (COMET and CometWiki, our best RemBERT-based multi-score model) against different translation quality dimensions on the test set.}
\label{tab:comet}
\end{table*}

\begin{table}[t!]
  \centering
%  \small
%\resizebox{\columnwidth}{!}{%
\begin{tabular}{@{}cccc@{}}
\toprule
\textbf{Model} & \textbf{Single} & \textbf{Multi} & \textbf{$\Delta$ Score}     \\ \midrule
RemBERT (MTE)    & 0.39                      & 0.42                     & +0.03 \\
mT5 (MTE)        & 0.38                      & 0.37                     & -0.01         \\
XLM-R (QE)       & 0.35                      & 0.38                     & +0.03 \\
RemBERT (QE)     & 0.39                      & 0.41                     & +0.02 \\ \bottomrule
\end{tabular}%
%}
\caption{Experiment 3: Correlations (Kendall's $\tau$) for multi-score and single-score models of overall translation quality}
\label{tab:exp3}
\end{table}

\subsection{Experiment 2: Impact of Training Data Size on Model
  Performance}

Experiment~2 varies the amounts of training data between 200 and 1000
training data points. Figure\ \ref{fig:exp2} shows the performance of
the four best models, per dimension and averaged. The rightmost
results (for 1000 training datapoints) correspond to the values in
Table~\ref{tab:exp1}. We see that even though the models obtain
reasonable level of performance for 1000 training datapoints, there
still appears to be a fairly linear improvement with increasing data
size, especially in the accuracy and fluency dimensions, indicating
that further improvements are possible with more training data.
Overall, the four models behave comparatively similar to one
another.

\subsection{Experiment 3: Predicting Overall Translation Quality}

Experiment 3 focuses on the prediction of overall translation
quality. It compares the results obtained when predicting overall
translation quality directly (``single-score'') vs. predicting the
three individual quality dimensions (``multi-task'') first and then
accumulating them to obtain the overall quality. Table\ \ref{tab:exp3}
shows the results, again for the four best models. For three of our
four models, the multi-task setting improves over the single-task
setting, and the positive effect in the three cases is more
substantial than the single negative case.  These findings highlight
the benefits of multi-score models in that they not only offer
fine-grained performance insights but also have the potential of
outperforming simpler models --- even if one is ultimately only
interested in a single overall quality assessment.

%\paragraph{Case Study: In-depth Translation Evaluation}
%As a qualitative analysis, we dive deeper into the predicted scores
%along with their input units: source, (reference) and hypothesis by
%the four selected models. We demonstrate their possible biases to
%certain error types by referring back to their annotations.
%\spa{Experiment 4? 5? Or attach to another experiment?}

\subsection{Experiment 4: Comparison against COMET}

While our models, trained as multi-output regressors, diverge from the
conventional approach of directly predicting overall quality scores,
we conducted a comparative analysis, contrasting our standout RemBERT
models with COMET-22 and CometKiwi, which are recognized as
state-of-the-art evaluators within the MTE and QE frameworks,
respectively. To ensure a fair comparison, we derive overall quality
predictions from COMET-22 and CometKiwi and evaluate them as before,
using Tau correlation, against both overall quality scores and the
MQM-derived specific quality dimensions. The results are detailed in
Table \ref{tab:comet}, alongside the findings from our RemBERT models
from above.

A notable observation from our analysis is that COMET’s scores, particularly COMET-22, are skewed towards accuracy. This contrasts with our multi-output regressor models, which demonstrate a more balanced correlation for accuracy and fluency, with style slightly behind. This disparity highlights that our approach to MT evaluation as multi-task learning can provide a reliable evaluation by integrating diverse aspects without losing focus on individual dimensions. 

Both COMET-22 and CometKiwi show a notably lower correlation for
fluency, approximately around 0.10, in contrast to our
models. While COMET-22’s correlation for style
diminishes further from its fluency score, CometKiwi demonstrates an
enhanced correlation with human-evaluated scores for style. This
suggests CometKiwi might better capture the nuances of style, aligning
with our findings in Experiment 1, where QE models generally
outperformed MTE models in evaluating style. Such an observation leads
us to speculate about the underlying mechanisms of QE models'
performance suggesting that QE models may act similarly to language
models in distinguishing between styles.

Finally, CometKiwi matches the performance of RemBERT (QE) when
predicting overall translation quality. This observation is noteworthy
since CometKiwi operates in a zero-shot setting for Korean, without
specific training on Korean MT data. Both CometKiwi and RemBERT have
undergone Korean pre-training. While CometKiwi was extensively
fine-tuned on a vast MT multlingual evaluation dataset (657k sentence
pairs), our RemBERT-based multi-score prediction model was fine-tuned
on 1k Korean sentences. This demonstrates an interesting trade-off
between focused language-specific and broad language-agnostic tuning.
%\spa{please double check}
%\dpa{Thank you for your refinement!}

%% 6. Discussions and Conclusions
\section{Discussion and Conclusions}
%\spa{Generally, discuss results as close as possible to the results
%  themselves, i.e. in the previous section. Only discuss results here
%  if (a) they arise from multiple experiments together or (b) if they
%  are important at the 'take-home message' level.}
%\spa{Make sure that the Conclusion lifts the discussion to a higher
%  level -- not just, what have we seen here? but also, what does it
%  mean for other people? why should they care? what can we do that we
%  could not do before? For this paper, discussing the results in terms
%  of explainability / attribution of ML models more generally might be
%  an interesting contextualization.}

This paper tackles a bottleneck for using the Multidimensional Quality
Metric (MQM) framework for automatic fine-grained MT, namely the
scarcity of suitably annotated resources. We introduce a benchmark
dataset for English--Korean, demonstrating the feasibility of an
annotation setup for three major error categories (accuracy, fluency,
style) and show its utility to train automatic models for MQM-based
three-dimensional MT evaluation models, reframing MT evaluation as
multi-task learning.

%x
% \dpa{I've added the phrase "by reframing MT evaluation as multi-task learning" to emphasize our method to deal with the task.}.
Empirically, we find that RemBERT consistently emerges as
the standout among other SOTA models in both MTE (reference-based) and
QE (reference-free) setups.% for English-Korean MT evaluation.
%In the detailed dimension-wise evaluation,
Our finding that reference-free models outperform their
counterparts in the style dimension, while reference-based models
excel in the accuracy dimension is in line with earlier
findings that the evaluation of style is more influenced by fidelity
to the original source than to the reference. In contrast, reference
translations still hold importance in determining factual
correctness and completeness
\cite{10.1162/tacl_a_00330,sun-etal-2020-estimating}. Furthermore, we
demonstrate that the multi-dimensional evaluation approach not only
enhances the interpretability of MT evaluation systems at the level of
fine-grained dimensions but also rivals or even surpasses single-score
models in terms of overall quality score. Our approach is also
competitive with the MT evaluation metrics of the Comet family, arguably
striking a better balance between different dimensions of translation quality.

Multi-dimensional evaluation is a strong desideratum for MT
practitioners to navigate trade-offs in translation
\citep{Alves2020TheRH, Lim2024SimpsonsPA}. For instance, scientific
reports prioritize adequacy, whereas literary works like novels or
poems require more fluent and idiomatic expressions, sometimes at the
expense of literal adequacy.
%These varied use cases underscore the
%importance of distinguishing between different quality aspects in
%translation.
By giving users access to evaluation at these different dimensions,
our approach yields enhanced explainability of translation quality
assessments and potentially contributes to better user acceptance.

\section*{Limitations}

%\spa{please write :) -- reuse text from conclusions about limitations?}
%\dpa{I've updated it by moving the text from conclusions and addressing two %notable concerns by the reviewers.}

A major limitation of our study is that we only consider a single
language pair, namely English-Korean. At the same time, this language
pair is known to be challenging due to the typological differences
between the two languages
\cite{hong-etal-2005-customizing,choi-etal-2018-automatic}. Therefore,
we take the good results we obtain regarding annotation reliability
and regarding automatic prediction quality to be promising, also with
respect to generalizing our approach to other language pairs.

While the weighting of individual dimensions in determining an overall
quality score may vary based on the specific translation objectives,
we adopted a straightforward method employing averages for
general-purpose evaluation, primarily to establish a simple evaluation
schema comparable to other MT evaluation setups
\citep{rei-etal-2022-comet, rei-etal-2022-cometkiwi}. Our method
however straightforwardly supports adjusting the weights to reflect
the relative significance of different translation quality aspects for
specific contexts of translation.

Another direction that we did not explore is the use of cross-lingual
transfer learning methods to address the need for manual annotation.
Future work can build on studies leveraging existing MT evaluation
datasets from other languages \citep{Freitag2021ExpertsEA} to address
this challenge, extending them from the overall quality case to the
multidimensional MQM case.

\section*{Ethics Statement}

In this study, we introduce a benchmark dataset for English-Korean
translation evaluation. We have ensured that all data utilized is
publicly available and does not contain any personal or identifiable
information. We involved several annotators in our primary annotation
and cross-validation process. All were informed about the purpose of
the research and the methods employed. The dataset and the model
implementation are publicly available.

We believe that our fine-grained automatic MT evaluation holds the
potential to enhance trust in neural MT systems by presenting results
in a more interpretable manner. As research, including our own,
continues to improve the interpretability of these neural systems, we
believe that such detailed evaluations can better position them to
handle translations that are sensitive to cultural and societal
contexts.

%We have conducted our study with constant honesty and transparency, ensuring accurate reporting of methods, data, and results. While our initial focus centers on English-Korean translation evaluation, this choice is based on the specific challenges of this language pair, not out of any bias towards specific languages. Our overarching goal is to understand the potential of pre-trained models, and we plan to expand our investigation to encompass a broader range of language pairs in the future.

\section{References}
\label{sec:reference}

%% Bibliographie
\bibliographystyle{lrec-coling2024-natbib}
\bibliography{lrec-coling2024-example}

%% Language Resource References
\section{Language Resource References}
\label{lr:ref}

\bibliographystylelanguageresource{lrec-coling2024-natbib}
\bibliographylanguageresource{languageresource}

% \end{document}

% Appendix
\clearpage
\appendix
\onecolumn

\section{Model Details}
\label{sec:model-details}
%The tables below list the properties of the LMs we use as well as
%the specific input encoding we employ.

\begin{table}[hb!]
\centering
\begin{tabular}{ccccccc}
\toprule
\textbf{Model} & \textbf{Params} & \textbf{Blocks} & \textbf{Heads} & \textbf{Emb Size} & \textbf{FFN Dim} & \textbf{Max Len} \\ \midrule
mBERT          & 178M            & 12              & 12             & 768                 & 3072             & 512              \\
XLM-R          & 660M            & 24              & 16             & 1024                & 4096             & 514              \\
RemBERT        & 576M            & 32              & 18             & 1152                & 4608             & 512              \\
mBART (Enc)    & 408M            & 12              & 16             & 1024                & 4096             & 1024             \\
mT5 (Enc)      & 564M            & 24              & 16             & 1024                & 2816             & 1024              \\
M2M100 (Enc)   & 635M            & 24              & 16             & 1024                & 8192             & 1024             \\ \bottomrule
\end{tabular}
\caption{Architectural specifics of the selected models in terms of model size, depth, and inherent complexity.}
\end{table}

\begin{table}[tbh!]
\centering
\begin{tabular}{@{}cll@{}}
\toprule
\multicolumn{1}{l}{Setup} & Model Type  & Token Format                                                                                                                                               \\ \midrule
\multirow{4}{*}{MTE}  & Encoder Models & \texttt{[CLS]} \textit{source} \texttt{[SEP]} \textit{reference} \texttt{[SEP]} \textit{hypothesis} \texttt{[SEP]}                                                       \\
                      & mBART          & \texttt{en\_XX} \textit{source} \texttt{ko\_KR} \textit{reference} \texttt{ko\_KR} \textit{hypothesis} \texttt{</s>}                                                      \\
                      & mT5            & \texttt{<init>} \textit{source} \texttt{<sep>} \textit{reference} \texttt{<sep>} \textit{hypothesis} \texttt{</s>}                                                      \\
                      & M2M100         & \texttt{\_\_en\_\_} \textit{source} \texttt{\_\_ko\_\_} \textit{reference} \texttt{\_\_ko\_\_} \textit{hypothesis} \texttt{</s>}                                      \\ \midrule
\multirow{4}{*}{QE}   & Encoder Models & \texttt{[CLS]} \textit{source} \texttt{[SEP]} \textit{hypothesis} \texttt{[SEP]}                                                                                           \\
                      & mBART          & \texttt{en\_XX} \textit{source} \texttt{ko\_KR} \textit{hypothesis} \texttt{</s>}                                                                                       \\
                      & mT5            & \texttt{<init>} \textit{source} \texttt{<sep>} \textit{hypothesis} \texttt{</s>}                                                                                      \\
                      & M2M100         & \texttt{\_\_en\_\_} \textit{source} \texttt{\_\_ko\_\_} \textit{hypothesis} \texttt{</s>}                                                                               \\ \bottomrule
\end{tabular}
\caption{Token Formats for Different Models Across Tasks.}
\label{tab:app:input-encoding}
\end{table}

\pagebreak

\section{Score and Error Distribution of the MQM Dataset}
\label{sec:score-error-distr}

%%score distribution
\begin{figure}[h]
  \centering
  \includegraphics[width=0.9\linewidth]{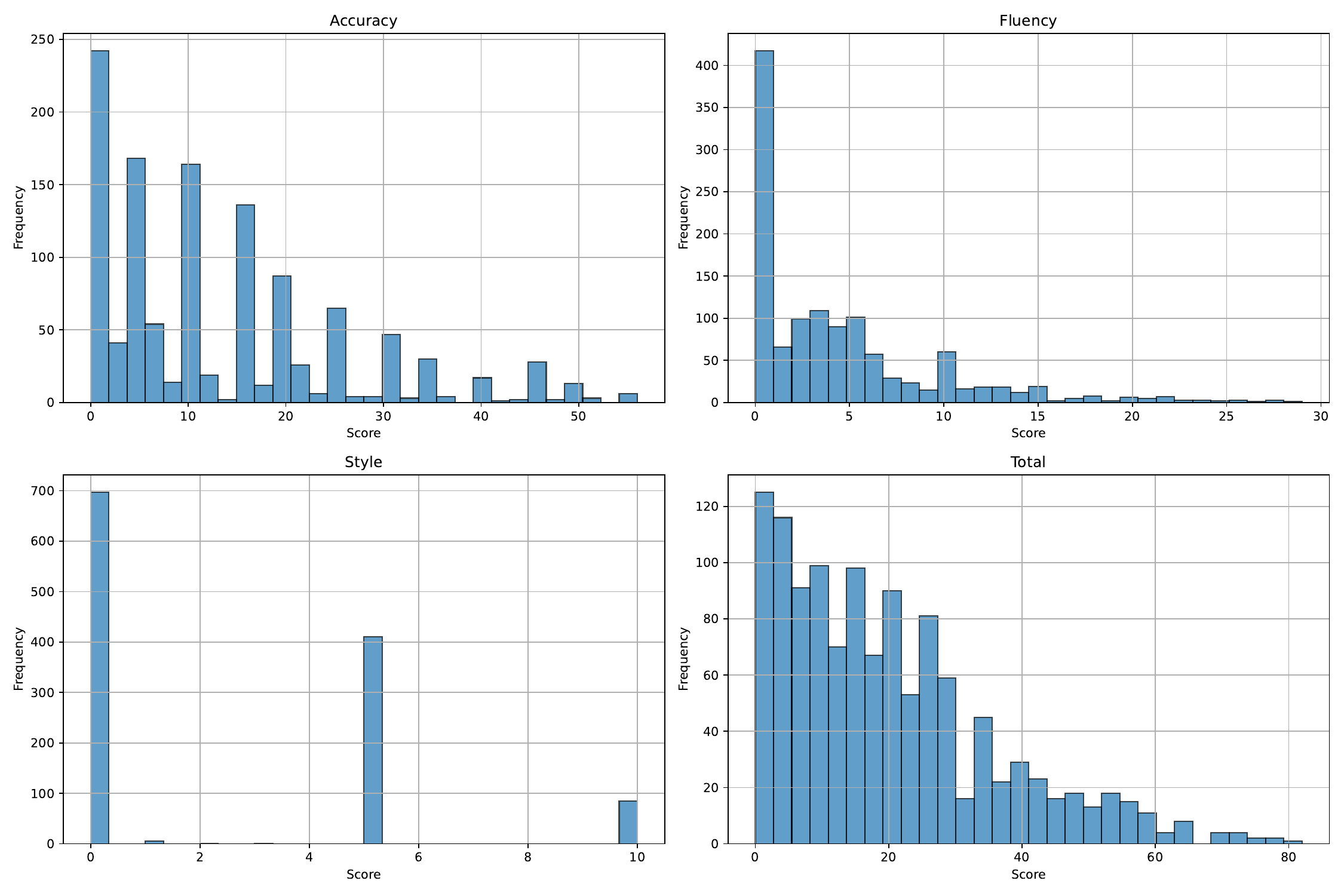}
  \caption{MQM Score Distribution by Dimension.}
\end{figure}

% \begin{figure*}[h]
%   \centering
%   \includegraphics[width=0.95\linewidth]{figures/score_dist_gv.pdf}
%   \caption{MQM Score Distribution by Dimension for Global Voices}
% \end{figure*}

% \begin{figure*}[h]
%   \centering
%   \includegraphics[width=0.95\linewidth]{figures/score_dist_ted.pdf}
%   \caption{MQM Score Distribution by Dimension for TED Talks 2020}
% \end{figure*}

\begin{figure}[h]
  \centering
  \includegraphics[width=0.9\linewidth]{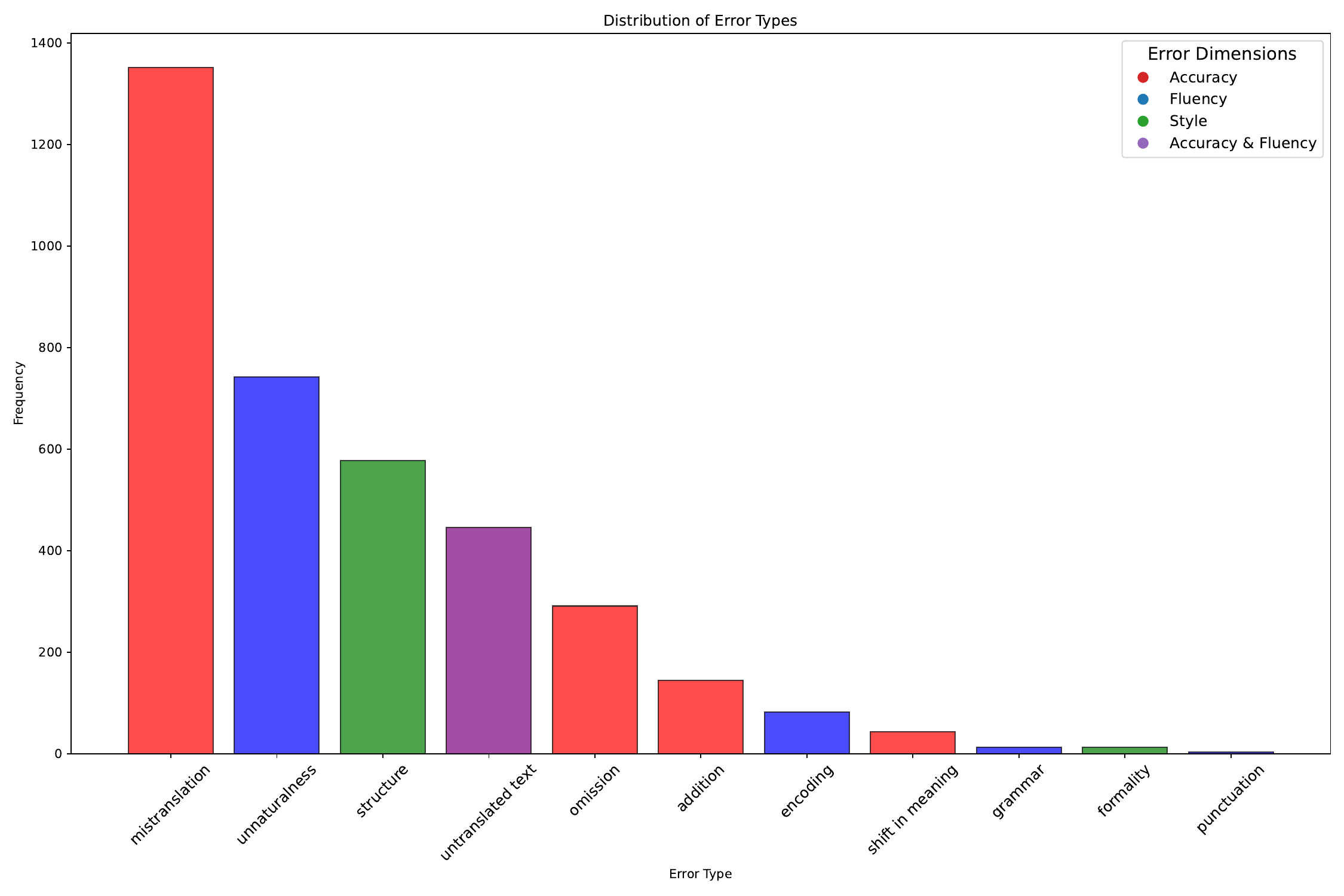}
  \caption{MQM Error Distribution by Dimension.}
\end{figure}

%%error distribution
% \begin{figure*}[h]
%   \centering
%   \includegraphics[width=0.95\linewidth]{figures/error_dist_gv.pdf}
%   \caption{MQM Error Distribution by Dimension for Global Voices}
% \end{figure*}

% \begin{figure*}[h]
%   \centering
%   \includegraphics[width=0.95\linewidth]{figures/error_dist_ted.pdf}
%   \caption{MQM Error Distribution by Dimension for TED Talks 2020}
% \end{figure*}

\pagebreak

\begin{CJK}{UTF8}{mj}
\section{Error Cases along with the Scores Predicted by our MTE/QE Models}
\label{sec:error-case}

\begin{table}[h]
\centering
\begin{tabular}{@{}>{\centering\arraybackslash}p{3cm}>{\centering\arraybackslash}p{3cm}>{\centering\arraybackslash}p{3.7cm}>{\centering\arraybackslash}p{1.5cm}>{\centering\arraybackslash}p{1.5cm}>{\centering\arraybackslash}p{1.7cm}@{}}
\toprule
{\centering\arraybackslash}\textbf{Source Text (EN)} & \textbf{Target Text (KO)} & \textbf{Annotation} & \textbf{Golden Score} & \textbf{RemBERT (MTE)} & \textbf{RemBERT (QE)} \\ \midrule
\parbox[t]{3.2cm}{His report matches this photograph shared by Kareem Fahim on Twitter, which shows Morsi supporters, carrying sticks and shields, and wearing helmets.} & \parbox[t]{3.2cm}{그가 제시한 보고서는 Kareem Fahim이 Twitter에서 공유한 사진과 일치합니다. 이미지는 헬멧을 쓰고 방패와 막대기를 들고 있는 무르시 지지자들을 묘사합니다.} & \parbox[t]{3.7cm}{Accuracy: Kareem \\Fahim이 Twitter\\(untranslated text/major) \\Fluency: Kareem \\Fahim이 Twitter\\(untranslated text/major) \\Style: -} & 30 & \textbf{34.16} & \textbf{30.04} \\ \midrule
\parbox[t]{3.2cm}{And when I got myself together and I looked at her, I realized, this isn't about me.} & \parbox[t]{3.2cm}{정신을 차리고 그녀를 관찰한 결과 상황이 내 중심으로 돌아가는 것이 아니라는 사실을 깨달았다.} & \parbox[t]{3.7cm}{Accuracy: 관찰한\\(mistranslation/minor), 상황이 내 중심으로 \\돌아가는\\(mistranslation/major)\\ Fluency: -\\ Style: -} & 21 & \textbf{21.14} & \textbf{18.83} \\ \bottomrule
\end{tabular}
\caption{Cases of overall score predictions; it demonstrates that predicted scores align closely with golden scores when only a single type of error is present.}
\end{table}

\begin{table}[h]
\centering
\begin{tabular}{@{}>{\centering\arraybackslash}p{3cm}>{\centering\arraybackslash}p{3cm}>{\centering\arraybackslash}p{3.7cm}>{\centering\arraybackslash}p{1.5cm}>{\centering\arraybackslash}p{1.5cm}>{\centering\arraybackslash}p{1.7cm}@{}}
\toprule
\textbf{Source Text (EN)} & \textbf{Target Text (KO)} & \textbf{Annotation} & \textbf{Golden Score} & \textbf{RemBERT (MTE)} & \textbf{RemBERT (QE)} \\ \midrule
\parbox[t]{3.2cm}{It's a coming of age for bringing data into the humanitarian world.} & \parbox[t]{3.2cm}{인도주의 분야는 데이터 통합의 중추적 순간을 경험하고 있습니다.} & \parbox[t]{3.7cm}{Accuracy: 분야는\\(mistranslation/major), \\중추적 순간을\\(mistranslation/major)\\ Fluency: -\\ Style: 인도주의 분야는\\ .. 경험하고 있습니다.\\(structure/major)} & 20 & \textbf{21.77} & \textbf{22.03} \\ \midrule
\parbox[t]{3.2cm}{We have babies.} & \parbox[t]{3.2cm}{우리 가족에게는 신생아가 있습니다.} & \parbox[t]{3.7cm}{Accuracy: 가족에게는\\(addition/major)\\ Fluency: -\\ Style: 우리 가족에게는 \\신생아가 있습니다.\\(structure/major)} & 10 & \textbf{10.90} & \textbf{8.49} \\ \bottomrule
\end{tabular}
\caption{Cases of overall score predictions; it showcases that predictions match closely with golden scores, especially in cases where sentences are notably short.}

\end{table}

\pagebreak

\begin{table}[h]
\centering
\begin{tabular}{@{}>{\centering\arraybackslash}p{3cm}>{\centering\arraybackslash}p{3cm}>{\centering\arraybackslash}p{3.7cm}>{\centering\arraybackslash}p{1.5cm}>{\centering\arraybackslash}p{1.5cm}>{\centering\arraybackslash}p{1.7cm}@{}}
\toprule
\textbf{Source Text (EN)} & \textbf{Target Text (KO)} & \textbf{Annotation} & \textbf{Golden Score} & \textbf{RemBERT (MTE)} & \textbf{RemBERT (QE)} \\ \midrule
\parbox[t]{3.2cm}{The occupy protests are not a color revolution.} & \parbox[t]{3.2cm}{점거 시위는 색깔에 기반한 혁명을 구성하지 않습니다.} & \parbox[t]{3.7cm}{Accuracy: 색깔에 \\기반한\\(mistranslation/minor), \\구성하지 않습니다.\\(mistranslation/major)\\ Fluency: 색깔에 기반한 \\혁명을 구성하지 \\않습니다.\\(unnaturalness/minor)\\ Style: -} & 12 & 7.77 & 6.08 \\ \midrule
\parbox[t]{3.2cm}{Does Guinea even have a blogosphere to speak of?} & \parbox[t]{3.2cm}{기니에 중요한 블로깅 커뮤니티가 있습니까?} & \parbox[t]{3.7cm}{Accuracy: 중요한\\(addtion/major), 블로깅\\(mistranslation/minor), \\to speak of\\(omission/major)\\ Fluency: -\\ Style: -} & 21 & 12.48 & 13.69 \\ \bottomrule
\end{tabular}
\caption{Cases of accuracy score predictions; models often face challenges in accurately predicting scores when encountering domain-specific terminology like
`color revolution' or `blogosphere'.}
\end{table}

\begin{table}[h]
\centering
\begin{tabular}{@{}>{\centering\arraybackslash}p{3cm}>{\centering\arraybackslash}p{3cm}>{\centering\arraybackslash}p{3.7cm}>{\centering\arraybackslash}p{1.5cm}>{\centering\arraybackslash}p{1.5cm}>{\centering\arraybackslash}p{1.7cm}@{}}
\toprule
\textbf{Source Text (EN)} & \textbf{Target Text (KO)} & \textbf{Annotation} & \textbf{Golden Score} & \textbf{RemBERT (MTE)} & \textbf{RemBERT (QE)} \\ \midrule
\parbox[t]{3.2cm}{His report matches this photograph shared by Kareem Fahim on Twitter, which shows Morsi supporters, carrying sticks and shields, and wearing helmets.} & \parbox[t]{3.2cm}{그가 제시한 보고서는 Kareem Fahim이 Twitter에서 공유한 사진과 일치합니다. 이미지는 헬멧을 쓰고 방패와 막대기를 들고 있는 무르시 지지자들을 묘사합니다.} & \parbox[t]{3.7cm}{Accuracy: Kareem \\Fahim이 Twitter\\(untranslated text/major)\\ Fluency: Kareem \\Fahim이 Twitter\\(untranslated text/major)\\ Style: -} & 15, 15 & \textbf{16.68, 14.15} & \textbf{13.1, 14.58} \\ \midrule
\parbox[t]{3.2cm}{The authorities have declared a state of emergency while the Prime Minister Mohammed Ghannouchi announced on state television that he was taking over as interim President.} & \parbox[t]{3.2cm}{모하메드 간누치(Mohammed Ghannouchi) 총리가 국영 TV에서 그가 임시 대통령의 역할을 맡을 것이라고 선언한 것과 함께 비상사태가 관리들에 의해 발표되었습니다.} & \parbox[t]{3.7cm}{Accuracy: (Mohammed \\Ghannouchi)\\(untranslated text/minor)\\ Fluency: (Mohammed \\Ghannouchi)\\(untranslated text/minor)\\ Style: 비상사태가 \\관리들에 의해 \\발표되었습니다.\\(structure/major)} & 2, 2 & 12.2, 8.76 & 12.01, 10.27 \\ \bottomrule
\end{tabular}
\caption{Cases of accuracy and fluency score predictions; minor errors tend to be overly penalized, especially in untranslated texts (bottom), contrastive to how major errors (top) are assessed.}
\end{table}

\pagebreak

\begin{table}[h]
\begin{tabular}{@{}>{\centering\arraybackslash}p{3cm}>{\centering\arraybackslash}p{3cm}>{\centering\arraybackslash}p{3.7cm}>{\centering\arraybackslash}p{1.5cm}>{\centering\arraybackslash}p{1.5cm}>{\centering\arraybackslash}p{1.7cm}@{}}
\toprule
\textbf{Source Text (EN)} & \textbf{Target Text (KO)} & \textbf{Annotation} & \textbf{Golden Score} & \textbf{RemBERT (MTE)} & \textbf{RemBERT (QE)} \\ \midrule
\parbox[t]{3.2cm}{According to the source, such a reading of the law was supported by mainland Chinese legal experts.} & \parbox[t]{3.2cm}{소식통에 따르면 중국 본토의 법률 전문가들은 이러한 법률 해석을 지지한 것으로 알려졌습니다.} & \parbox[t]{3.7cm}{Accuracy: the\\(omission/major)\\ Fluency: -\\ Style: 중국 본토의 법률 \\전문가들은 이러한 법률 \\해석을 지지한 것으로 \\알려졌습니다.\\(structure/major)} & 5 & \textbf{6.03} & \textbf{2.94} \\ \midrule
\parbox[t]{3.2cm}{If you think about what being a great parent is, what do you want? What makes a great parent?} & \parbox[t]{3.2cm}{뛰어난 부모의 자질을 고려하십시오. 귀하의 이상적인 기준은 무엇입니까? 육아에서 위대함을 구성하는 것은 무엇입니까?} & \parbox[t]{3.7cm}{Accuracy: If\\(omission/major), \\육아에서 위대함을 \\구성하는\\(mistranslation/major)\\ Fluency: 고려하십시오.\\(unnaturalness/minor), \\육아에서 위대함을 \\구성하는 것은\\(unnaturalness/minor)\\ Style: 뛰어난 부모의 \\자질을 고려하십시오.\\(structure/major), \\육아에서 위대함을 \\구성하는 것은 \\무엇입니까?\\(structure/major)} & 20 & \textbf{20.53} & \textbf{26.94} \\ \midrule
\parbox[t]{3.2cm}{Below is a brief explanation of the landmarks with photos taken by Au Kalun, a former journalist and a famous blogger.} & \parbox[t]{3.2cm}{저명한 블로거이자 전 저널리스트인 Au Kalun은 주목할만한 랜드마크에 대한 간결한 설명과 함께 시각 자료를 제공했습니다.} & \parbox[t]{3.7cm}{Accuracy: Below\\(omission/major), \\주목할만한\\(addition/major)\\ Fluency: Au Kalun\\(untranslated text\\/major)\\ Style: 시각 자료를 \\제공했습니다.\\(structure/major)} & 10 & 22.94 & 18.56 \\ \bottomrule
\end{tabular}
\caption{Cases of accuracy score predictions; predictions tend to match closely with the golden scores when frequent words such as `the' or `if' (top) are excluded, unlike the case with `below', a deictic term (bottom).}
\end{table}

\end{CJK}

\pagebreak

\pagebreak

\section{MQM Annotation Guidelines for English-Korean Translation}
\label{sec:annot-guid}
\begin{CJK}{UTF8}{mj}
\subsection{Introduction}

These guidelines have been created to streamline the evaluation process of English-to-Korean translation quality, aligning with the Multidimensional Quality Metrics (MQM) framework. The MQM framework provides a robust structure enabling evaluators to judge the quality of translation under uniform and clear-cut criteria, thereby converting the abstract concept of translation quality into measurable values.

Evaluation of translation quality is intrinsically prone to subjectivity, with individual evaluators often holding different standards for what constitutes a good or poor translation. This variability can lead to significant discrepancies in the assessment results across different evaluators. The MQM framework, therefore, is utilized to provide explicit error classification criteria to mitigate such inconsistencies. It facilitates the conversion of these classifications into scores based on a uniform set of standards, leading to a more objective and comparable translation quality assessment.

\subsection{Overview of the MQM Evaluation Process}

The evaluation of translation quality consists of two distinct phases: ``error annotation'' and ``score conversion''. In this task, your primary focus will be on ``error annotation''. The ``Score Conversion'' phase, while not part of this task, is explained here for a more comprehensive understanding.

During the ``error annotation'' task, your job is to evaluate the quality of translation at the level of individual translation units, following the MQM framework. Each unit consists of original English text and its Korean translation. Your role is to critically compare and analyze both the original and translated texts in each unit, identifying and annotating any evident errors. Words that are deemed erroneous should be annotated according to their sub-error type and severity level under the respective error dimensions. This process is based on three main error dimensions: accuracy, fluency, and style. Please note that the error dimension of terminology, outlined in the official MQM, is not included in this task due to the non-domain-specific nature of the corpora used for our evaluation (news and presentations). More details on the annotation will be provided in section 4, ``Annotation Process''.

The subsequent ``score conversion'' phase involves turning the annotated errors into quantifiable scores that reflect the translation quality. Major errors are assigned a weight of 5 points, while minor errors are given 1 point. The total scores for all errors within each error dimension are then added up to create the MQM dimension score. The MQM total score is obtained by aggregating these MQM dimension scores by these three error dimensions.

To maintain the integrity of this process, it is crucial that you adhere strictly to the definitions and instructions provided in the following sections during the annotation task. The definitions are refined based on the official MQM specifications \footnote{https://themqm.org/}.

\subsection{Error Dimensions}

\subsubsection{Accuracy}

This dimension evaluates whether the original meaning is well conveyed in the translated text. Words that change the original meaning are considered accuracy errors. The sub-error types under accuracy include:

\begin{itemize}
    \item Addition: Inserting words that are not present in the original text.
    \item Omission: Leaving out words from the original text.
    \item Shift in meaning: Placing words in a different part of the text than originally intended, resulting in a change in meaning.
    \item Mistranslation: Translating words from the original text into words with different meaning.
    \item Untranslated text: Leaving words from the original text untranslated.
\end{itemize}

The severity of errors within the accuracy dimension is classified as major or minor. Major errors significantly distort the overall meaning of the text, whereas minor errors do not considerably impede comprehension. For example, if the English word ``cell phone'' is translated as ``전자 기기''(electric device) and the main idea of the original message can still be conveyed, it belongs to minor errors. However, if it is translated as ``사무 용품''(office supplies) and the original message is significantly impacted, then it should be categorized as a major error.

For the error type ``untranslated text'', any untranslated English words in the translation are considered major errors. However, untranslated words in parentheses alongside their Korean translation, such as ``페이스북''(Facebook), are deemed minor errors. Hashtags, such as \#MeToo, are viewed as loanwords and should be left untranslated. Therefore, translated hashtags are considered mistranslation errors.

After analyzing the original and translated text, errors are generally defined within the translated text. However, for omission errors, errors should be identified in the original text since they cannot be located in the translation.

\subsubsection{Fluency}

This dimension evaluates whether the translated text reads smoothly and naturally. Any words that disrupt the readability of the translated text are considered as fluency errors. Unlike other error dimensions, fluency is evaluated solely based on the translated text, not the original text. The sub-error types under fluency include:

\begin{itemize}
    \item Grammar: Violating the grammar rules of the target language.
    \item Spelling: Words that are misspelled.
    \item Punctuation: Using punctuation incorrectly (e.g., commas, periods, question marks, exclamation marks, quotation marks, etc.).
    \item Encoding: Misrepresentations due to incorrect encoding processes (e.g., ``\&quot;'' appearing where a quotation mark '' should be).
    \item Formatting: Violating conventional formats required in a particular region.
    \item Unnaturalness: Words that are awkward or unnatural.
    \item Untranslated text: Leaving words from the original text untranslated.
\end{itemize}

The severity of errors within the fluency dimension is also categorized as major and minor. Major errors significantly hinder the readability of the text, while minor errors, though not obstructive to understanding, can reduce the text's overall quality.

Taking into account that 'unnaturalness' errors have generally a minor impact on translation quality compared to other sub-error types such as grammar or encoding, they are typically classified as minor errors.

Untranslated text is considered an error in both fluency and accuracy since these errors not only impact the meaning of the text but also disrupt its readability. These errors are identified using the same criteria as explained in accuracy: untranslated words in parentheses alongside their Korean translation are considered minor errors, while words left entirely untranslated are considered major errors.

In the official MQM specification, formatting is classified under a separate error dimension, 'local convention'. However, given the rarity of such errors in this evaluation, formatting errors have been incorporated as a sub-error type under the fluency dimension. Formatting errors include incorrect representation of time, date, and currency formats. For instance, while the date format in the United States is MM-DD-YYYY, it is YYYY-MM-DD in Korea. Translations should mirror these local conventions.

Please note that fluency errors do not consider the meaning of the original text but do take into account the semantic relationship within the translated text. Hence, if a text's flow feels unnatural due to incorrect semantic connections between words, this is deemed an 'unnaturalness' error.

\subsubsection{Style}

This dimension evaluates whether the original writing style is adequately preserved in the translated text. Any words or phrases that deviate from the style of the original text are considered style errors. The sub-error types under style include:

\begin{itemize}
    \item Formality: Having differences in formality between the original and the translated text, or inconsistent formality within the translated text.
    \item Structure: Having structural changes that affect the nuances of the original text (e.g., order of writing, passive/active voice, or word/sentence conversion).
\end{itemize}

The severity of errors within the style dimension is categorized as major or minor, depending on the degree of influence the error has on the overall tone of the text. Major errors are those that have a clear impact on the overall tone of the text. For example, shifting from a formal to an informal tone in the translated text, or modifying the text structure in a way that changes the original nuances. Minor errors, on the other hand, are those that have a minimal impact on the overall tone of the text. These may include slight modifications in sentence structure or minor fluctuations in formality.

Style errors can span across the entire text and are often not limited to specific words. Therefore, while other error types are calculated based on word count, style errors are calculated per text unit. In this context, a text unit refers to a sequence of words that is found to be erroneous. For instance, if a sequence like ``어떤 영향을 미칠지 불확실했습니다'' (translated as ``it was uncertain what influence it would have'') is deemed to cause a style shift, this text unit is counted as a single error, not four. This distinguishes style errors from other error dimensions, as they focus on the overall style of the text, which can affect an entire sentence or multiple phrases.

\subsection{Error Annotation Process}

This section describes the annotation process. During this task, it's important to stay strictly within the context of the given text pairs. Words or phrases that exceed the provided context should be annotated as errors. The annotation will be performed according to the three error dimensions, accuracy, fluency, and style which are explained in section 3.

Error annotation is to be carried out sequentially, one dimension at a time. If no errors are found within the dimension under examination, type a hyphen ``-''. This practice differentiates between error dimensions that are yet to be annotated and those without errors.

It's important to remember that a single word can potentially fall into multiple error dimensions simultaneously. For example, a word could be annotated as an error for both accuracy and fluency. However, it's not possible for the same word to be categorized under different sub-error types within the same dimension, such as grammar and unnaturalness in the fluency dimension. In cases where this might occur, the more severe error is chosen for annotation. The layout for each translation unit to be evaluated is as follows:
\\\\
\noindent [n-th translation unit]\\
\noindent \textit{Original English Text}\\
\noindent \textit{Translated Korean Text}\\\\
\noindent Accuracy: \\
\noindent Fluency: \\
\noindent Style:\\

Your task is to identify errors within the original text and its translation and annotate them under their respective accuracy, fluency, and style dimensions. Each error should be annotated in the following format: `Error\_Word\_or\_Phrase(Sub\_Error\_
Type/Severity\_Level)'. Here's an example:
\\\\
\noindent [n-th translation unit]\\
\noindent And demonstrations also occurred in Ni'lin. \\
\noindent Ni'lin은 또한 시위가 일어나는 것을 목격했습니다.\\\\
\noindent Accuracy: Ni'lin(untranslated text/major), And(omission/minor), 목격했습니다.(mistranslation/major) \\
\noindent Fluency: Ni'lin(untranslated text/major), 또한(unnaturalness/minor) \\
\noindent Style: Ni'lin은 또한 시위가 일어나는 것을 목격했습니다.(structure/major)\\

The Korean translation “Ni'lin은 또한 시위가 일어나는 것을 목격했습니다.” translates to “Ni'lin also witnessed the occurrence of demonstrations”. Firstly, in terms of accuracy, “Ni'lin” remains untranslated in the translation, significantly impacting the translation quality. Hence, it's annotated as a major error under “untranslated text”. The word “And”, which should have been translated to “그리고” in Korean, is missing from the translated text, but this omission is considered to have a minor impact on the overall meaning of the translation. Therefore, it's annotated as a minor error under “omission”. “목격했습니다”(witnessed) is an incorrect translation that distorts the original meaning. If it were translated as “경험했습니다”(experienced), it could have preserved the original text's meaning. For that reason, it's annotated as a major error under “mistranslation”.

Next, in terms of fluency, the untranslated word “Ni'lin” significantly affects the text's readability, so it's annotated as a major error under the “untranslated text”. While “또한” is the correct Korean word for its counterpart “also”, the Korean translation doesn't sound natural with it, so it's annotated as a minor error under the “unnaturalness”.

Lastly, the sentence structure has changed from active to passive voice, and this shift noticeably impacts the overall tone of the translation. Therefore, the text unit “Ni'lin은 또한 시위가 일어나는 것을 목격했습니다.” where this structural change occurs is annotated under the style dimension as a major error.

After your annotation work is done, the MQM scores can be calculated in the next score conversion phase. The annotation example above would yield MQM dimension scores of 11, 6, and 5 for accuracy, fluency, and style, respectively, resulting in an MQM total score of 22.

\subsection{Final Remarks}

Whenever you face uncertainty, don't hesitate to revisit these guidelines. Your deep understanding and correct implementation of the MQM framework directly influence the quality and consistency of the evaluations. Remember, your meticulous work is crucial in maintaining high standards of translation quality evaluation. Thank you for taking the time to read these guidelines and good luck with your annotation work!

\end{CJK}

\end{document}